\documentclass{amsart}
\usepackage{hyperref}
\usepackage{xr}
\usepackage{amsfonts}
\usepackage[T1]{fontenc}
\usepackage{enumitem}
\usepackage[margin=4cm]{geometry}
\usepackage{amsmath,amssymb,amsfonts,amsthm}
\usepackage{xfrac}
\usepackage{mathtools}
\usepackage{bm}
\usepackage{latexsym}
\usepackage{amscd}
\usepackage[all,cmtip]{xy}
\usepackage{mathrsfs}
\usepackage{upgreek}
\usepackage{tikz-cd}
\usepackage{graphicx}
\usepackage{enumitem} 
\usepackage{braket}
\usepackage{verbatim}
\usepackage{stmaryrd}
\usepackage{cancel}

\allowdisplaybreaks

\numberwithin{equation}{section}

\theoremstyle{plain}
\newtheorem{proposition}{Proposition}[section]

\theoremstyle{definition}

\newtheorem{remarks}[proposition]{Remark}

\theoremstyle{plain}

\theoremstyle{remark}

\usepackage{ifthen}    

\makeatletter

\newcommand{\textfrac@kern}{%
  \ifx\textfrac@nexttoken\@sptoken%
  \else%
    \kern.08333em%
  \fi%
}

\newcommand{\textfrac@nostar}[3][]{%
  \mbox{%
    \ifthenelse{\not\equal{#1}{}}
      {#1\/\kern.05em}
      {}
    \raisebox{.775ex}{\tiny #2}
    \raisebox{.365ex}{\kern-.15em{\scriptsize /}\kern-.15em}
    \raisebox{0ex}{\tiny #3}
  }%
  \futurelet\textfrac@nexttoken\textfrac@kern%
}

\newlength\textfrac@width@num%
\newlength\textfrac@width@denom%
\newlength\textfrac@width@%
\newcommand{\textfrac@star}[3][]{%
  \settowidth{\textfrac@width@num}{\tiny #2\/}%
  \settowidth{\textfrac@width@denom}{\tiny #3\/}%
  \ifthenelse{\lengthtest{\textfrac@width@num<\textfrac@width@denom}}%
    {\let\textfrac@width@\textfrac@width@denom}%
    {\let\textfrac@width@\textfrac@width@num}%
  \mbox{%
    \ifthenelse{\not\equal{#1}{}}
      {#1\/\kern.08333em}
      {}
    \ooalign{%
      \relax\cr%
      \noalign{\vskip-1.1ex}%
      {\hss\tiny #2\/\hss}\cr
      \noalign{\vskip1.1ex}%
      \rule[.6666ex]{\textfrac@width@}{.4pt}\cr
      \noalign{\vskip.4ex}%
      {\hss\tiny #3\/\hss}\cr
      \noalign{\vskip-.4ex}%
    }%
  }%
  \let\textfrac@width\undefined%
  \futurelet\textfrac@nexttoken\textfrac@kern%
}

\def\textfrac{\@ifstar\textfrac@star\textfrac@nostar}

\makeatother

\makeatletter
\renewcommand\part{%
   \if@noskipsec \leavevmode \fi
   \par
   \addvspace{4ex}%
   \@afterindentfalse
   \secdef\@part\@spart}

\def\@part[#1]#2{%
    \ifnum \c@secnumdepth >\m@ne
      \refstepcounter{part}%
      \addcontentsline{toc}{part}{\thepart\hspace{1em}#1}%
    \else
      \addcontentsline{toc}{part}{#1}%
    \fi
    {\parindent \z@ \raggedright
     \interlinepenalty \@M
     \normalfont
     \ifnum \c@secnumdepth >\m@ne
       \Large\textsc \partname\nobreakspace\thepart
       \par\nobreak
     \fi
     \huge \textsc #2%
     \par}%
    \nobreak
    \vskip 3ex
    \@afterheading}
\def\@spart#1{%
    {\parindent \z@ \raggedright
     \interlinepenalty \@M
     \normalfont
     \huge \textsc #1\par}%
     \nobreak
     \vskip 3ex
     \@afterheading}
\makeatother

\newcommand{\PP}{\mathbb{P}}

\newcommand{\RR}{\mathbf{R}}

\newcommand{\cA}{\mathcal A}
\newcommand{\cB}{\mathcal B}

\newcommand{\cE}{\mathcal E}

\newcommand{\cN}{\mathcal N}

\newcommand{\cT}{\mathcal T}

\newcommand{\cV}{\mathcal V}

\def\tn{\mathtt{n}}

\newcommand{\beq}{\begin{equation}}
\newcommand{\eeq}{\end{equation}}
\newcommand{\beqa}{\begin{eqnarray*}}
\newcommand{\eeqa}{\end{eqnarray*}}

\def\multiset#1#2{\ensuremath{\left(\kern-.3em\left(\genfrac{}{}{0pt}{}{#1}{#2}\right)\kern-.3em\right)}}   

\def\wl{\par \vspace{\baselineskip}}

\newcommand{\nc}{n_{\mathrm{ctx}}}
\newcommand{\nv}{n_{\mathrm{vocab}}}

\renewcommand{\dh}{d_{\mathrm{h}}}

\DeclareMathOperator{\sm}{softmax}
\DeclareMathOperator{\Attn}{\textsc{attn}}
\DeclareMathOperator{\FF}{\textsc{ff}}

\author{Spencer Becker-Kahn}

\title[Mathematics of GPT LLM Architectures]{Notes on the Mathematical Structure of GPT LLM Architectures}

\begin{document}

\maketitle

\section*{Introduction}

When considered from a purely mathematical point of view, the building and training of a large (transformer) language model (LLM) is the construction of a function - which can be taken to be a map from some euclidean space to another - that has certain interesting properties. And therefore, from the point of view of a mathematician, it may be frustrating to find that many key papers announcing significant new LLMs seem reluctant to simply spell out the details of the function that they have constructed in plain mathematical language or indeed even in complete pseudo-code (and the latter form of this complaint appears to be one of the motivations behind a recent article of Phuong and Hutter \cite{phuonghutterformal}). Here, we seek to give a relatively `pure' mathematical description of the architecture of a GPT-3-style LLM.


 \subsection*{Trainable Parameters}
 
 Like all such models in machine learning, the construction really initially describes a family of functions indexed by some set $\Theta = \mathbf{R}^{N}$ called the \emph{parameter space}. There is then a separate process - the training of the model - in which a particular value $\theta \in \Theta$ is selected using a training algorithm. Each dimension of $\Theta$ corresponds to the possible values of an individual \emph{trainable parameter}. We will draw attention to such parameters as we introduce them, as opposed to attempting to give a definition of $\Theta$ up front. But this short note discusses only the architecture and does not describe any training algorithms.
 
\section{Tokenization, Encodings and Embeddings} \label{S:Tok}

 In this section we describe the constituent parts of the mechanism by which a transformer language model handles real text.

\subsection{The Vocabulary}  

Let $\mathscr{C}$ denote a corpus of real text: This will typically be a large collection of different artefacts (articles, books, blog posts, text conversations etc.) from many different places on the internet. We will use this to define a \emph{vocabulary} for the transformer. To begin this process, we will first think of the corpus as one (very long) string $\alpha_1\alpha_2\alpha_3\dots = (\alpha_i)_{i \in I}$ of characters. To do this, a \emph{special character} e.g. `$\star$' will have been introduced, which separates the individual artefacts in the corpus, i.e. the character `$\star$' marks where one thing (i.e. an article/book/post) ends and the next bit of text begins. Every other character in the corpus is drawn from a fixed standard alphabet e.g. $\mathcal{A} = \{a,b,c,\dots,A,B,C,\dots,!,@,\%,\dots\}$. 

\wl

\noindent \textbf{The Base Vocabulary.} The vocabulary $\cV$ will be constructed via an iterative process, starting with what we will call the base vocabulary, which we will now describe. Recall that a \emph{byte} is an element of $B = \{0,1\}^8$, i.e. a byte is a string of 8 bits. The UTF-8 encoding $\mathcal{E}_{\mathrm{UTF-8}}$ is a standard way to map elements of $\mathcal{A}$ to short strings of bytes. It is a variable-length encoding: The strings can have length between 1 and 4, i.e. the encoding can be thought of as an injective map 
\begin{align}
\mathcal{E}_{\mathrm{UTF-8}} : \mathcal{A}\ \longrightarrow\ B \cup (B \times  B) \cup (B \times  B \times B) \cup (B \times  B \times B\times B).
	\end{align}
The \emph{base vocabulary} $\cV_0$ is then the set of symbols consisting of the byte strings $\cE_{\mathrm{UTF-8}}(\cA)$ together with two special characters to denote the ends of words and ends of individual artefacts. 

Next, by replacing every character $\alpha$ in the corpus $\mathscr{C}$ by its image $\mathcal{E}_{\mathrm{UTF-8}}(\alpha)$ under the map $\mathcal{E}_{\mathrm{UTF-8}}$,  we produce a new corpus $\mathscr{C}_0$ (or rather a `translation' of the original corpus) that is written entirely using symbols from the base vocabulary $\mathcal{V}_0$. 

\wl

\noindent \textbf{Byte Pair Encoding.} Next, we produce $\mathscr{C}_1$ and $\mathcal{V}_1$ from $\mathscr{C}_0$ and $\mathcal{V}_0$ in the following way: We take the pair of symbols $(s,s')$ that occurs the greatest number of times consecutively in $\mathscr{C}_0$ (excluding special characters) and add the single new symbol $\tilde{s} = ss'$ to the vocabulary $\mathcal{V}_0$, i.e. we set $\mathcal{V}_1 = \mathcal{V}_0 \cup \{ ss' \}$. And then we replace every occurrence of $ss'$ in the corpus $\mathscr{C}_0$ by the new symbol $\tilde{s}$, resulting in the corpus $\mathscr{C}_1$. At this stage we also define the \emph{merge rule} $\mathcal{M}_1 := (ss',\tilde{s})$ to be the ordered pair that `records' the fact that the symbols $s$ and $s'$ were merged into the symbol $\tilde{s}$ when they appear consecutively.

We then continue iteratively, producing $\mathcal{V}_{i+1}$ and $\mathscr{C}_{i+1}$ from $\mathcal{V}_i$ and $\mathscr{C}_i$ in the same way, and producing $\mathcal{M}_{i+1}$ at each such stage too. Notice that the size of the vocabulary is increased by one at each step, i.e. $|\mathcal{V}_{i+1}| = |\mathcal{V}_i| + 1$.  We fix a parameter $\nv$ and terminate the process at $\mathcal{V} := \mathcal{V}_m$ with $|\mathcal{V}_m| = \nv$. We refer to $\mathcal{V}$ as \emph{the vocabulary}.  The base vocabulary $\mathcal{V}_0$ together with all of the merge rules $(\mathcal{M}_i)_{i=1}^m$, forms the \emph{byte pair encoding} $\mathcal{B} := \bigl( \mathcal{V}_0, (\mathcal{M}_i)_{i=1}^m\bigr)$. An element of the vocabulary is called a \emph{token}. 

 The result is: 

\begin{enumerate}[label=\textbf{\arabic*$\dagger$)}, ref=\textbf{\arabic*$\dagger$)}]
\item \label{1dagger} (Tokenization) By using $\mathcal{E}_{\mathrm{UTF-8}}$ and $\cB$, we have a canonical way of taking any string $S$ of text (that is written using the symbols from the alphabet $\cA$) and mapping it to a finite sequence of tokens (i.e. elements from the fixed vocabulary $\mathcal{V}$).
\end{enumerate}

\begin{remarks} Two remarks are in order:

\indent 1) The way we have described the construction of the vocabulary via byte pair encoding would mean going through the entire corpus multiple times, but in practice this process would be done on a much smaller sample taken from the full corpus. 

\indent 2) Given a token $v \in \cV$, if we consider the `un-merged' version - i.e. write the token using symbols from $\cV_0$ - one can then act on each (non-special character) symbol using $\cE^{-1}_{\mathrm{UTF-8}}$ to produce a string written in the alphabet $\cA$. In practice it is usually these strings that people are referring to when they use the word `token'. For example, one may say that the fragment `ed' is a token.
\end{remarks}

\noindent \textbf{One-Hot Encoding.} Next, we fix what is called a \emph{one-hot encoding} of the vocabulary, which is simply a bijection between the set $\mathcal{V}$ and the standard orthonormal basis of $\mathbf{R}^{\nv}$. Denote this by $\sigma : \mathcal{V} \to  \{e_j\}_{j=1}^{\nv}$.  One can refer to $\sigma(\mathcal{V}) \subset \RR^{\nv}$ as the set of \emph{one-hot tokens}.
\begin{enumerate}[resume*]
\item \label{2dagger} (One-hot encoding) By using $\mathcal{E}_{\mathrm{UTF-8}}$, $\cB$, and $\sigma$, we have a canonical way of taking any string $S$ of text (that is written using the symbols from the alphabet $\cA$) and mapping it to a finite sequence $ (t_1(S),t_2(S),\dots,t_N(S))$ of one-hot vectors in $\RR^{\nv}$, where $N = N(S)$.
\end{enumerate}
Note that we will re-use the notation introduced here, i.e. that $N(S)$ denotes the number of tokens that the string $S$ is split into after tokenization.

\subsection{Embedding and Unembedding}\label{SS:Embedding}
The next steps are about continuing to repackage the sequential language data into a format that is more suitable for deep learning. First, we set a very important parameter that will be the height of the matrices that the main part of the transformer takes as its inputs. It is called the \emph{size of the context window} and we will denote it by $\nc$ or just $n$ when there is no ambiguity. So,
\begin{enumerate}[resume*]
\item \label{3dagger} Using $\cE_{\mathrm{UTF-8}}$, $\cB$, and $\sigma$, we now have a canonical way of taking any string $S$ of text that is written using the symbols from the alphabet $\cA$ and which satisfies $N(S) = n$ and forming a matrix $t(S) \in \RR^{\nc \times \nv}$, the rows of which are the one-hot tokens $t_1(S),t_2(S),\dots,t_n(S)$.
\end{enumerate}
We may refer to a string $S$ with $N(S) = \nc$ as \emph{the context}.  Next, we embed the one-hot encoding in a smaller vector space. Specifically, we choose a parameter $d < \nv$ and a $(d \times \nv)$ projection matrix
\begin{align}
W_E : \RR^{\nv} \rightarrow\ \RR^{d}
\end{align}
called the \emph{token embedding}. The entries of the matrix $W_E$ are trainable parameters (and after training an embedding like this is referred to as a \emph{learned embedding}). We refer to $W_E(\sigma(\mathcal{V}))$ as the set of \emph{embedded tokens}. Notice that since the one-hot tokens are literally an orthonormal basis of $\RR^{\nv}$, we have that the embedded tokens are the columns of the embedding matrix $W_E$. 

Now, given a matrix $t$ of one-hot tokens, as described in \ref{3dagger}, the first thing that the transformer does is act on each row of $t$ by the embedding matrix, \emph{i.e.}
\begin{align}
t \xmapsto[\mathrm{Embedding}]{} (\mathrm{Id} \otimes W_E) t = t W_E^T = X =   \underbrace{\left( \begin{matrix} [\ \rule[0.5ex]{1.2cm}{1pt}\ x_1\ \rule[0.5ex]{1.2cm}{1pt}\ ]  \\     \vdots    \\  [\ \rule[0.5ex]{1.2cm}{1pt}\ x_i\ \rule[0.5ex]{1.2cm}{1pt}\ ]   \\ \vdots \\  [\ \rule[0.5ex]{1.2cm}{1pt}\ x_n\ \rule[0.5ex]{1.2cm}{1pt}\ ] \end{matrix} \right)}_{d}  \left. \begin{matrix} 
 \vphantom{} \\ \vphantom{} \\ \vphantom{} \\ \vphantom{} \\ \vphantom{}  \\ \vphantom{}   \end{matrix} 
 \right\} \nc
\end{align}
We also define a $(\nv \times d)$ matrix
 \begin{align}
W_U : \RR^{d} \rightarrow\ \RR^{\nv}
\end{align}
called the \emph{unembedding}, the entries of which are also trainable parameters. This acts on the rows of a matrix $X \in \RR^{n \times d}$, i.e.
\begin{align}
X \xmapsto[\mathrm{Unembedding}]{} (\mathrm{Id} \otimes W_U) X = X W_U^T = 
  \underbrace{\left( \begin{matrix} [\ \rule[0.5ex]{2cm}{1pt}\ \tau_1\ \rule[0.5ex]{2cm}{1pt}\ ]  \\     \vdots    \\  [\ \rule[0.5ex]{2cm}{1pt}\ \tau_i\ \rule[0.5ex]{2cm}{1pt}\ ]   \\ \vdots \\  [\ \rule[0.5ex]{2cm}{1pt}\ \tau_n\ \rule[0.5ex]{2cm}{1pt}\ ] \end{matrix} \right)}_{\nv}  
 \left. \begin{matrix} 
 \vphantom{} \\ \vphantom{} \\ \vphantom{} \\ \vphantom{} \\ \vphantom{}  \\ \vphantom{}   \end{matrix} 
 \right\} \nc
\end{align}
Sometimes one insists that $W_U = W_E^T$, a constraint that we refer to as the embedding and unembedding being `\emph{tied}'.

\begin{remarks} One also typically includes some kind of \emph{positional encoding}.  There are several competing ways of doing this which we will not review here and this is one of a few aspects of these models that we essentially just ignore in these notes. Let us just say that the simplest positional encoding is just to define another $n \times d$ matrix $P$ the entries of which are all trainable parameters and to simply \emph{add it on} to $X$, i.e. during the embedding step, the transformer maps $t \mapsto tW_E^T + P$.	We have opted to leave out any further discussion of positional encodings.
\end{remarks}

\section{Feedforward Layers}

We will describe the multi-layer perceptron by first describing a generalized feedforward architecture.  We will not need to use these architectures in the full generality presented here, but it does serve to illustrate the underlying mathematical structure at the level of the individual vertices or `aritificial neurons' better than would be achieved by directly describing the multi-layer perceptron in the way it is usually thought of. 

\subsection{Basic General Definitions} \label{SS:BasicGeneralDef}
Given a directed, acyclic graph $G = (V,E)$ with minimum degree at least one (i.e. with the property that every vertex belongs to at least one edge), we write $V = I \cup H \cup O$ where:
\begin{itemize}
		\item  $I$ is the set of vertices which have no incoming edges; these are called the \emph{input} vertices.
		\item  $O$ is the set of vertices, which have no outgoing edges; these are called the \emph{output} vertices.
		\item $H := V \setminus (I \cup O)$.
	\end{itemize}
	
A \emph{feed-forward artificial neural network} $\mathcal{N}$ is a pair $(G,A)$ where 
\begin{enumerate}
\item $G = (V,E)$ is a finite, directed acyclic graph with minimum degree at least one called the \emph{architecture} of $\mathcal{N}$; and
\item $A = \{ \sigma_v : \RR \to \RR : v \in H\}$ is a family of functions called the \emph{activation functions} for $\mathcal{N}$.
\end{enumerate}
In a feed-forward artificial neural network, the input and output vertices are labelled, so that we may write $I = \{I_1,\dots,I_d\}$, where $d = |I|$, and $O = \{O_1,\dots,O_{d'}\}$, where $d' = |O|$.  

Given a choice of \emph{weights} $w_e \in \RR$ for $e \in E$ and \emph{biases} $b_v \in \RR$ for $v \in V\setminus I$, the network defines a function $m : \RR^d \to \RR^{d'}$ in the following way (for now we have omitted from the notation the dependence of $m$ on $\mathcal{N}$, $(w_e)_{e \in E}$, and $(b_v)_{v \in V\setminus I}$): For $x \in \RR$ and $j=1,\dots,d$, define
\[
z_{I_j}(x) := x_j.
\]
Then, for any $v \in V\setminus I$, the \emph{preactivation} at $v$ is given by
\begin{align} \label{E:preactivation_at_v}
z_v(x) := b_v + \sum_{e = (v',v) \in E} w_e \sigma_{v'}(z_{v'}(x)),
\end{align}
and the \emph{activation} at $v$ is given by:
\begin{align} \label{E:activation_at_v}
a_v(x) := \sigma_v\bigl( z_v(x) \bigr).
\end{align}
Thus
\begin{align} \label{E:activations}
a_v(x) = \sigma_v\Bigl( b_v + \sum_{e=(v'v) \in E} w_e a_{v'}(x)\Bigr).
\end{align}
The output of the function $m$ is given by the preactivations at the output vertices, i.e.  $m(x) := \bigl(z_{O_1}(x),\dots, z_{O_{d'}}(x)\bigr)$.

The \emph{parameter space} of the network $\mathcal{N}$ is a set $\Theta_{\mathcal{N}}$, each point of which represents a choice of weights and biases, i.e. $\Theta_{\cN}$ is the collection of all  $\theta = (w,b)$ where $w = (w_e)_{e \in E} \in \mathbf{R}$ is a set of weights for the network and $b = (b_v)_{v \in V\setminus I} \in \mathbf{R}$ is a set of biases. The weights and biases are referred to as the trainable parameters of the network.

\subsection{The Fully Connected Multi-Layer Perceptron}

The \emph{multilayer perceptron (MLP)} is a feed-forward artificial neural network for which:
\begin{enumerate}
\item The architecture $G = (V,E)$ is \emph{layered}: This means that 
\item The vertex set $V$ can be written as the disjoint union $V = V^{(0)} \cup \dots \cup V^{(L)}$, where 
\begin{enumerate}
\item $L \geq 1$ called the \emph{depth}\footnote{This is the `deep' in \emph{deep learning}} of the network;
\item $I= V^{(0)}$ and $O = V^{(L)}$; and 
\item $E \subset \bigcup_{l=0}^{L-1} V^{(l)}\times V^{(l+1)}$; 
\end{enumerate}
\item The activation functions are all equal to the same function $\sigma$.
\end{enumerate} 
In these notes we will be concerned only with MLPs that are \emph{fully connected} which means that 
\begin{align} \label{E:fullyconnected}
E &= \bigcup_{l=0}^{L-1} V^{(l)}\times V^{(l+1)}.
\end{align}
The layers $V^{(1)},\dots, V^{(L-1)}$ are called the \emph{hidden layers}. The integer $n_l := |V^{(l)}|$ is called the \emph{width} of the $l^{th}$ layer and we write $v^{(l)}_i$ for the $i^{th}$ vertex in the $l^{th}$ layer of the network.  Given $\theta = (w,b) \in \Theta_{\cN}$, write $b^{(l)}_i = b_{v^{(l)}_i}$ so that $b^{(l)}$ is a vector containing all the biases at the $l^{th}$ layer of the network and for $l=1,\dots,L$ let $W^{(l)} \in \RR^{n_{l}\times n_{l-1}}$ denote a matrix whose entries are given by
\[
w^{(l)}_{ij}  := w_{(v^{(l)}_i,  v^{(l-1)}_j)}.
\]
This is called a weight matrix.

The special structure of the MLP means that it is fruitful to describe it in terms of how it maps one layer to the next, as opposed to using a description that stays only at the level of the individual vertices.  Given $x \in \RR^d$,  define the $l^{th}$ \emph{layer preactivation} to be the vector $z^{(l)}(x) \in \RR^{n_l}$ whose components are given by 
\[
z^{(l)}_i(x) := z_{v^{(l)}_i}(x).
\]
And define the \emph{$l^{th}$ layer activation} to be the vector $a^{(l)}(x) \in \RR^{n_l}$ given by 
\[
a^{(l)}_i(x) = a^{v^{(l)}_i}(x).
\]
Then \eqref{E:preactivation_at_v} implies that
\begin{align} \label{E:z^l}
z^{(l+1)}(x) =  b^{(l+1)} + W^{(l+1)} \sigma\bigl(z^{(l)}(x)\bigr),
\end{align}
where in this equation and henceforth we we will use the convention that the activation function $\sigma : \RR \to \RR$ acts component-wise when applied to a vector (and here of course $W^{(l+1)}$ acts by matrix multiplication on $\sigma\bigl(z^{(l)}(x)\bigr)$),  i.e.  if we were to write out the components in full,  then we would have:
\begin{align} \label{E:z^l_i}
z^{(l+1)}_i(x) = b^{(l+1)}_i + \sum_{j=1}^{n_l} w^{(l+1)}_{ij}\sigma\bigl( z^{(l)}_j(x)\bigr)
\end{align}
for $i = 1,\dots,n_{l+1}$ and $l=0,\dots,L-1$.  And \eqref{E:activations} implies that
\begin{align}
a^{(l+1)}(x) = \sigma \Bigl( b^{(l+1)} +  W^{(l+1)} a^{(l)}(x)\Bigr).
\end{align}
So for the multilayer perceptron, the map which takes the activations at a given layer as inputs and then outputs the activations at the next layer is an affine transformation followed by the application of the activation function component-wise.  In this context, the space $\RR^{n_{l+1}}$  - that which contains the $(l+1)^{th}$ layer activations $a^{(l+1)}(x)$ - is called \emph{activation space}. And recall that the function that the MLP implements is given by the preactivations at the final layer, i.e. $m(x) = z^{(L)}(x)$.

\begin{remarks} One may actually want to implement activations at the final layer, but it seems better to leave the definition like this since it is easy to consider $\sigma \circ m$ if one needs to. Also note that often in theoretical work, one needn't bother with biases at all; they can be simulated by instead creating an additional input dimension at each layer which only ever receives the input `$1$'. Then the action of adding a bias can be replicated via a weight on the new input edge. ( It's not clear to me this is really any simpler than just leaving the biases there but it explains why you sometimes don't see them. )\end{remarks}

\subsection{Feedforward Layers} Fully-connected MLPs are incorporated into the transformer architecture via feedforward layers. With $n,d \geq 1$ fixed, given a matrix $X \in \RR^{n \times d}$, and a fully-connected MLP $m$ with input dimension and output dimensions equal to $d$ (i.e. with $n_0 = n_L = d$), we will write $m(X)$ to denote the $(n\times d)$ matrix obtained by letting $m$ act independently on each row of $X$.  And given such an MLP we define a \emph{feedforward layer} $\FF_m$ by
\begin{align*}
\FF_m : \mathbb{R}^{n\times d} &\longrightarrow \mathbb{R}^{n\times d} \\
X &\longmapsto X + m(X).
\end{align*} 
 
 \begin{remarks} In practice, it is common in transformer LLMs for the MLPs in the feedforward layers to have only one hidden layer, i.e. to just have $L=2$, though there is no real reason to insist on this in our theoretical description.  
\end{remarks}
 
\section{Attention Layers}

In this section we will describe the action of attention layers. One of the key structural difference between attention layers and feedforward layers is that whereas a feedforward layer processes each row of an input matrix $X \in \RR^{n\times d}$ independently, an attention layer performs operations `across' the rows of the matrix.

\subsection{Softmax with Autoregressive Masking}

With $n \geq 1$ fixed, define the softmax function 
\[
\sm : \mathbb{R}^{n\times n} \to \mathbb{R}^{n\times n}
\]
to act on the matrix $A = [a_{ij}]$ via the formula
\[
\sm(A)_{ij} = \frac{e^{a_{ij}}}{\sum_{p,q} e^{a_{pq}}}.
\]
We will write $\mathrm{softmax}^*$ for a modified version of the softmax function that uses what is known as \emph{autoregressive masking}.  The modified formula is:
\[
\sm^*(A)_{ij} = \begin{cases} 0 & \text{if}\ \ i < j \\  \frac{e^{a_{ij}}}{\sum_{p \geq q} e^{a_{pq}}} & \text{else}.\end{cases}
\]
We can think of this as being given by $\mathrm{softmax}^*(A) = \mathrm{softmax}(A + M)$, where $M$ is an $(n\times n)$ matrix called a \emph{mask} of the form
\[
m_{ij} = \begin{cases} 0 & \ \text{if}\ i \geq j \\
-\infty &\ \text{else} \end{cases}
\]
i.e.
\begin{align*}
M = \begin{pmatrix} 0 & -\infty &  \dots & \dots & -\infty \\
0 & 0 & -\infty & \dots & \vdots \\
0 & \cdots & 0  & -\infty & \vdots \\
\vdots & \cdots & \ddots & 0 & -\infty \\ 
0 & \cdots & \cdots & \cdots & 0 
\end{pmatrix}. 
\end{align*}

\subsection{Attention Heads}
An \emph{attention head} is a function
\begin{align}
h : \RR^{n\times d} \to \RR^{n\times d}
\end{align}
of the form
\begin{align}\label{E:attentionhead1}
h(X) = \biggl(\sm^*\bigl(X\,W_{\textsc{qk}}^h\, X^T \bigr) \otimes W_{\textsc{ov}}^h\biggr) X,
\end{align}
where $W_{\textsc{qk}}^h$ is a $d\times d$ matrix called the \emph{query-key} matrix and where $W_{\textsc{ov}}^h$ is a $d\times d$ matrix called the \emph{output-value} matrix. The entries of the query-key matrix and of the output-value matrix are all trainable parameters. Moreover both matrices are constrained to be products of two low-rank projections: There is another fixed integer $\dh < d$ called the \emph{dimension} of the attention head, and there are four projection matrices: 
\[
W_{\textsc{q}}^h , W_{\textsc{k}}^h, W_{\textsc{v}}^h : \RR^{d} \to \RR^{\dh},
\]
and 
\[
W_{\textsc{o}}^h : \RR^{\dh} \to \RR^{d}
\]
which are called the \emph{query}, \emph{key}, \emph{value} and \emph{output} matrices respectively, and for which 
\[
W_{\textsc{qk}}^h = (W_{\textsc{q}}^h)^TW^h_{\textsc{k}}
\]
and 
\[
W_{\textsc{ov}}^h := W_{\textsc{o}}^hW_{\textsc{v}}^h.
\]
So, as linear maps:
\begin{align}
W_{\textsc{qk}} : \RR^{d} \xrightarrow{\qquad W^h_K \qquad} \RR^{\dh} \xrightarrow{\qquad (W^h_Q)^T \qquad} \RR^{d}	
\end{align}
and
\begin{align}
W_{\textsc{ov}} : \RR^{d} \xrightarrow{\qquad W^h_V \qquad} \RR^{\dh} \xrightarrow{\qquad W^h_O \qquad} \RR^{d}.
\end{align}

\subsection{Attention Patterns}
The \emph{attention pattern} of the head $h$ is the function $A^h :\  \RR^{n\times d} \to \RR^{n\times n}$ given by
\begin{align}
A^h(X) = \sm^*\bigl(X\,W_{\textsc{qk}}^h\, X^T \bigr).
\end{align}
Notice that the expression $X W^h_{\textsc{QK}} X^T$ can be thought of as the result of applying a bilinear form to each pair of rows in the matrix $X$, \emph{i.e.} if we write $x_1,\dots, x_n \in \RR^d$ for the rows of $X$, and write $\langle \vec{v},\vec{v}'\rangle_{h} := \vec{v}W^h_{\textsc{QK}}\vec{v}'^T$ for any two vectors $\vec{v},\ \vec{v}' \in \RR^d$, then the attention pattern is
\begin{align*}
A^h\left(\left( \begin{matrix} [\ \rule[0.5ex]{1.2cm}{1pt}\ x_1\ \rule[0.5ex]{1.2cm}{1pt}\ ]  \\     \vdots    \\  [\ \rule[0.5ex]{1.2cm}{1pt}\ x_i\ \rule[0.5ex]{1.2cm}{1pt}\ ]   \\ \vdots \\  [\ \rule[0.5ex]{1.2cm}{1pt}\ x_n\ \rule[0.5ex]{1.2cm}{1pt}\ ] \end{matrix} \right) \right) := \sm^*\left( \begin{matrix}\langle x_1,x_1 \rangle_h &  \langle x_1,x_2 \rangle_h & \cdots & \langle x_1,x_n \rangle_h \\
\langle x_2,x_1 \rangle_h & \ddots & \vdots & \vdots \\
\vdots & & & \\
\langle x_n,x_1 \rangle_h & \cdots & \cdots & \langle x_n,x_n \rangle_h
\end{matrix} \right).	
\end{align*}
And note that once we take into account the notation of attention patterns and the way that the tensor product acts, we have
\begin{align} \label{E:attentionhead2}
h(X) = A^h(X)\ X\ W_{\textsc{ov}}^{h\ T}.
\end{align}

\subsection{Attention Layers}
An \emph{attention layer} is defined via an an \emph{attention multi-head}, which is a set $H$ of attention heads all with the same dimension,  i.e.  such that there is some integer $d_H$ with $\dh = d_H$ for every $h \in H$. Given such an attention multi-head, we define an attention layer $\Attn_H$ by
\begin{align*}
\Attn_H :\ \mathbb{R}^{n \times d} &\longrightarrow \mathbb{R}^{n \times d}\\
	X &\longmapsto X  + \sum_{h \in H}h(X).
\end{align*}

\begin{remarks} (Composition of Attention Heads) The product of two (or more) attention heads behaves in a similar way to a single true attention head via the fact that
\[
\bigl(A^{h_1}\otimes W^{h_1}_{\textsc{ov}}\bigr)\bigl(A^{h_2}\otimes W^{h_2}_{\textsc{ov}}\bigr) = A^{h_1}A^{h_2}\otimes W^{h_1}_{\textsc{ov}}W^{h_2}_{\textsc{ov}}.
\]
Thus the product behaves like an attention head with attention pattern given by $A^{h_1\circ h_2} = A^{h_1}A^{h_2}$ and with output-value matrix given by $W_{\textsc{ov}}^{h_1\circ h_2} = W_{\textsc{ov}}^{h_1}W_{\textsc{ov}}^{h_2}$. This is making use of the fact that for a fixed attention pattern, attention heads have a linear nature.
\end{remarks}

\section{The Full Transformer}

\subsection{Residual Blocks} Given a set $H$ of attention heads and a fully-connected MLP $m$, a \emph{residual block} $B = B(H,m) : \mathbb{R}^{n \times d} \to \mathbb{R}^{n \times d}$ is defined to be the composition of the attention layer $\Attn_H$ with the feedforward layer $\FF_m$, i.e.
\begin{align}
B(H,m) = \FF_m \circ  \Attn_H.
\end{align}
Recalling the definitions
\begin{align*}
\Attn_H &:  X   \longmapsto   X + \sum_{h \in H}h(X);\ \text{and} \\
\FF_m &: X  \longmapsto    X + m\bigl(X\bigr),
\end{align*}
we of course have
\begin{align}
B(X) = X + m\Bigl( X +  \sum_{h \in H}h(X) \Bigr).
\end{align}

\begin{remarks}
Some modern architectures choose to do these operations in `parallel' and instead use blocks that compute $X \mapsto X + \FF(X) + \Attn(X)$, but here we will stick with the composition which is used by, for example, the GPT-3 architecture.
\end{remarks}

\subsection{The Decoder Stack} With $n,d \geq 1$ fixed, fix another integer $\tn$, to be the number of blocks in the transformer. Let $\{H_i\}_{i=1}^{\tn}$ be a collection of attention multi-heads with $|H_i| = n_{\mathrm{heads}}$ for $i = 1,\dots, \tn$ and with $d_{H_i} = d_{\mathrm{heads}}$ for $i = 1,\dots, \tn$.  And let $\{m_i\}_{i=1}^{\tn}$ be a set of MLPs, each of which has $L$ layers and input and output dimensions equal to $d$.  We will refer to the composition $\mathcal{D}$ of the $\tn$ blocks $\{B(H_i,m_i)\}_{i=1}^{\tn}$ as the `\emph{decoder stack}' i.e.:
\[	
\mathcal{D}\ :\  \RR^{n\times d} \ \ \xrightarrow{B(H_1,m_1)} \ \  \RR^{n\times d}  \ \ \xrightarrow{B(H_2,m_2)}  \ \cdots\ \xrightarrow{B(H_{\tn-1},m_{\tn-1})} \ \ \RR^{n\times d}   \xrightarrow{B(H_{\tn},m_{\tn})}  \ \ \RR^{n\times d}.
\]
This function is the main part of the transformer that fits between the embedding and unembedding described in Section \ref{S:Tok}, as we will now explain: Given a matrix $t \in \mathbb{R}^{\nc \times \nv}$ - the rows of which are one-hot tokens as described in \ref{3dagger} of Section \ref{SS:Embedding} - the transformer first acts on $t$ via the embedding, then via the decoder stack, and then finally via the unembedding. So we define
\begin{align*}
\cT \ :\ \RR^{\nc \times \nv}\ \xrightarrow[\text{Embedding}]{Id\ \otimes\ W_E}\   \RR^{\nc \times d}\ \xrightarrow[\text{Decoder}]{\mathcal{D}}\  \RR^{\nc \times d}\ \xrightarrow[\text{Unembedding}]{Id\ \otimes\ W_U}\ \RR^{\nc \times \nv}.
\end{align*}
In this way, the transformer is a map:
\begin{align*}
\cT \ :\ \RR^{\nc \times \nv}\ \rightarrow\ \RR^{\nc \times \nv}.
\end{align*}

\subsection{Logits and Predictions}
Finally we will briefly describe how to interpret the output of the transformer as a prediction about the `next token'.  We will use the $\sm$ function on vectors in $\RR^{\nv}$, i.e. $\sm : \RR^{\nv} \longrightarrow\ \RR^{\nv}$, where for $x \in \RR^{\nv}$, the $i^{th}$ entry of $\sm(x)$ is given by
\begin{align*}
(\sm(x))_i = \frac{\exp(x_i)}{ \sum_{j=1}^{\nv} \exp(x_j)}.
\end{align*}
Of course,  we have $(\sm(x))_i \in [0,1]$ for every $i$ and $\sum_{i=1}^n (\sm(x))_i = 1$. Thus $\sm(x)$ defines a probability distribution on $\{1,\dots,\nv\}$. Now,  suppose we have a string $S$ of text for which $N(S) = \nc$ (recall that $N(S)$ denotes the number of tokens that the string $S$ is split into after tokenization; see Section \ref{S:Tok}). We form a matrix $t(S) \in \RR^{\nc \times \nv}$ (as explained in \ref{3dagger}). Then,  let $l(S) \in \RR^{\nv}$ denote the final row of the matrix $\cT\bigl(t(S)\bigr)$.  We refer to $l(S)$ as the \emph{logits} corresponding to the next token prediction for the context $S$.  By the 'next token', we mean the token that may come immediately after the end of the string $S$.

Finally we can define 
\[
\PP_S(i) := \bigl(\sm\bigl(l(S)\bigr)\bigr)_i.
\]
This gives a probability distribution on $\{1,\dots,\nv\}$ which has the interpretation that $\PP_S(i)$ is the probability that the model assigns to the possibility that the next token is the $i^{th}$ token in its vocabulary (recall that there is a bijection $\sigma : \mathcal{V} \to  \{e_j\}_{j=1}^{\nv}$ via which we consider the vocabulary $\cV$ to be ordered). In particular, $\arg \max_i \PP_S(i)$ can be interpreted as the index of the token that the model `believes' ought to be the next token.  Predicting one token at a time like this forms the basis of a transformer language model's ability to write in long form answers, though there are many more parts to the story from this point before one gets something that can write at the level of Chat GPT-type models (or Claude etc.).

\bibliographystyle{plain}
\bibliography{FHIbib}

\end{document}